\documentclass[letterpaper, 10 pt, conference]{ieeeconf}  

\IEEEoverridecommandlockouts                              

\usepackage{graphicx}
\usepackage{amsmath}
\usepackage{amssymb}
\usepackage{hyperref}
\usepackage{multirow}
\usepackage{float}
\usepackage{booktabs}
\usepackage{cite}
\usepackage{color}
\usepackage[ruled,vlined,linesnumbered]{algorithm2e}
\usepackage{algpseudocode}
\usepackage{placeins}
\usepackage{array}
\usepackage{listings}
\usepackage{caption}
\usepackage{subcaption}

\title{\LARGE \bf
TERL: Large-Scale Multi-Target Encirclement Using Transformer-Enhanced Reinforcement Learning
}

\author{Heng Zhang$^{1}$, Guoxiang Zhao$^{2,*}$ and Xiaoqiang Ren$^{1}$
    \thanks{The work was supported in part by the National Natural Science Foundation of China under Grants 62273223,  62336005, 62421004, 62403300, and 62461160313,
        the Shanghai Municipal Commission of Education under Grant 24SG38 and the Aeronautical Science Fund of China under Grant 2024Z0710S6001.}
    \thanks{$^{1}$Heng Zhang and Xiaoqiang Ren are with the School of Mechatronic Engineering and Automation, Shanghai University, Shanghai, China,
        {\tt\small \{heng-zhang,xqren\}@shu.edu.cn}.}%
    \thanks{$^{2}$Guoxiang Zhao is with the School of Future Technology, Shanghai University, Shanghai, China,
        {\tt\small gxzhao@shu.edu.cn}.}%
    \thanks{$^{*}$Corresponding author.}
}

\begin{document}

\maketitle
\thispagestyle{empty}
\pagestyle{empty}

    \begin{abstract}

        Pursuit-evasion (PE) problem is a critical challenge in multi-robot systems (MRS).
While reinforcement learning (RL) has shown its promise in addressing PE tasks, research has primarily focused on single-target pursuit, with limited exploration of multi-target encirclement, particularly in large-scale settings.
This paper proposes a \underline{T}ransformer-\underline{E}nhanced \underline{R}einforcement \underline{L}earning (TERL) framework for large-scale multi-target encirclement.
By integrating a transformer-based policy network with target selection, TERL enables robots to adaptively prioritize targets and safely coordinate robots.
Results show that TERL outperforms existing RL-based methods in terms of encirclement success rate and task completion time, while maintaining good performance in large-scale scenarios.
Notably, TERL, trained on small-scale scenarios (15 pursuers, 4 targets), generalizes effectively to large-scale settings (80 pursuers, 20 targets) without retraining, achieving a 100\% success rate.
The code and demonstration video are available at \url{https://github.com/ApricityZ/TERL}.

    \end{abstract}


    \section{INTRODUCTION}
    The complexity of many real-world tasks often exceeds the capabilities of a single robotic system, necessitating the deployment of multi-robot systems (MRS) to enhance efficiency, robustness, and scalability~\cite{rizk2020cooperative}.
As robotic applications expand, large-scale MRS is becoming essential for industrial automation, environmental monitoring, and infrastructure maintenance~\cite{yang2018grand, dorigo2021swarm, khan2020graph}.
Among these challenges, the PE problem~\cite{hajek2008pursuit}, where multiple pursuers coordinate to track and capture evasive targets, is a research focus, with applications in surveillance and patrol, containment of adversarial targets~\cite{wang2023spatio,hafez2014uavs}.

Conventional PE approaches, including differential game theory~\cite{weintraub2020introduction, li2024optimal} and heuristic methods~\cite{wang2022pursuit, li2017dynamics}, have been widely studied.
Differential game theory provides mathematically well-verified optimal strategies under proper assumptions about system dynamics and agent interactions.
Heuristic methods, including those that simulate force fields among entities, offer computational efficiency in various settings.
However, both approaches face significant challenges in complex real-world scenarios, particularly in dynamic, partially observable, and uncertain environments, and struggle to facilitate effective cooperative behaviors.

Recent works have explored Deep Reinforcement Learning (DRL)~\cite{mnih2015human} for PE problems, leveraging its adaptability to uncertain and dynamic environments.
With carefully designed policy networks and reward structures, DRL enables robots to cooperate and learn effective strategies for PE tasks without requiring explicit prior knowledge of the environment.
While DRL has demonstrated promising capabilities, existing research has predominantly focused on specific problem settings, leaving several critical aspects underexplored.
Most approaches~\cite{de2021decentralized, zhang2022multi, zhang2022game, wang2020cooperative} emphasize single-target \textit{pursuit}, where success is typically defined as a pursuer reaching a predefined distance from an evader.

However, pursuit does not fully capture the complexity of real-world scenarios, where coordinated encirclement or containment is often required~\cite{thummalapeta2023survey}.
Encirclement differs from pursuit in that the objective is to establish a containment formation instead of reaching a target position.
Furthermore, due to the absence of target selection mechanisms and encirclement formation guidance, single-target pursuit strategies do not naturally extend to multi-target encirclement, which introduces challenges beyond simple interception, particularly in large-scale scenarios.
Large-scale deployment involving numerous pursuers and evaders operating simultaneously requires not only increasing the number of robots but also ensuring safe coordination.
Existing methods typically concatenate or stack all received observations without explicitly modeling the relational information contained in them, thereby overlooking the interactions among diverse entities in the environment.

The key bottleneck limiting safety and generalization in large-scale multi-target encirclement lies in effectively modeling the heterogeneous, scale-dependent interactions among pursuers, evaders, and obstacles, especially in coordinated encirclement, target selection, and collision avoidance.
Transformer~\cite{vaswani2017attention} has demonstrated its remarkable effectiveness in capturing dependencies within long sequences and handling variable-length inputs.
It has been widely adopted in large-scale domains such as natural language processing~\cite{devlin2019bert}, where it models contextual dependencies among tokens, and computer vision~\cite{dosovitskiy2020image}, where it captures spatial relationships among image patches.
These paradigms exemplify the transformer's ability to extract structured relational patterns from complex, variable-sized inputs.
Such modeling capability closely parallels the challenges in multi-robot systems, where coordinated behavior depends on reasoning over dynamic interactions among heterogeneous entities, including pursuers, evaders, teammates, and obstacles.

This paper investigates the problem of multi-robot multi-target encirclement. We propose the \underline{T}ransformer-\underline{E}nhanced \underline{R}einforcement \underline{L}earning (TERL) framework, which integrates explicit relational reasoning into the network architecture and operates within the distributional reinforcement learning paradigm~\cite{bellemare2017distributional}.
Numerical experiments demonstrate its execution efficiency and safely commands large scale pursuers to encircle multiple moving targets.
Additionally, we provide open-source code and demonstration videos at \url{https://github.com/ApricityZ/TERL}.
The key contributions of this work are as follows:
1) This paper proposes a method to address the challenge of large-scale multi-target encirclement with collision-aware coordination in pursuit-evasion scenarios.
2) We propose a transformer-based relational reasoning policy network that facilitates dynamic target selection while enhancing cooperation and safety.
3) This study verifies the capability of the proposed network in handling large-scale multi-target encirclement, particularly in its mission completion rate and execution optimality and safety.

    \section{Related Work}
    
\textbf{Conventional PE approaches}.
The PE problem has attracted significant attention and has been studied using various approaches, including game theory, heuristic methods, and learning-based methods~\cite{mu2023survey}.
Among conventional approaches, differential game theory~\cite{weintraub2020introduction, li2024optimal} and heuristic strategies~\cite{wang2022pursuit, li2017dynamics} are widely adopted.
Differential game theory integrates optimal control and game theory to derive pursuit-evasion strategies, providing mathematically well-founded solutions that guarantee optimality under specific assumptions.
It has been extensively applied in adversarial settings where system dynamics and agent behaviors are well-defined.
However, its reliance on known system dynamics limits its applicability in uncertain and complex environments.
Heuristic approaches, often inspired by biological swarm behaviors~\cite{li2017dynamics}, use computational simulations to model agent interactions, offering flexibility in unknown environments.
Yet, their dependence on manually designed rules and empirical tuning hinders adaptability in dynamic, large-scale settings.

\textbf{RL for PE.}
Deep reinforcement learning has been widely adopted for PE problems, demonstrating significant advancements in decision-making and multi-agent coordination~\cite{de2021decentralized, zhang2022multi, zhang2022game, yang2023large, wang2020cooperative}.
Single-target pursuit has been extensively studied in~\cite{de2021decentralized, zhang2022multi, zhang2022game}.
Single-target encirclement, which focuses on coordination mechanisms to constrain the evader's movement, has been investigated in~\cite{wang2020cooperative}.
In multi-target scenarios, a strategy in which 24 pursuers pursued 6 evaders was examined in~\cite{yang2023large}.
These studies collectively demonstrate the effectiveness of learning-based methods in PE tasks.
Unlike conventional PE approaches that require both manually defined objectives and explicit strategy design,
reinforcement learning in general learns desired behaviors following a uniform framework of interactions with the environment and only requires a reward function that characterizes the satisfaction of planning objectives.
However, large-scale multi-target encirclement has received minimal attention, leaving a significant gap in the development of scalable and cooperative strategies for constraining multiple evaders in dynamic environments.

\textbf{Distributional RL}.
Distributional RL~\cite{bellemare2017distributional} models return distributions rather than single expected values, allowing agents to capture reward uncertainty and make risk-aware decisions.
By enhancing policy robustness and reducing collision risk, it improves robot safety~\cite{lin2024decentralized, liu2023adaptive}.
These advantages highlight the potential of distributional RL in safety-critical applications.

\textbf{Transformer in RL}.
Transformer~\cite{vaswani2017attention}, originally designed for sequence modeling, has demonstrated remarkable success in natural language processing~\cite{devlin2019bert} and computer vision~\cite{dosovitskiy2020image}, leading to increased interest in its application to reinforcement learning,
particularly in decision-making~\cite{chen2021decision} and multi-agent coordination~\cite{wen2022multi}.
The self-attention mechanism dynamically assigns importance to different entities in a sequence, allowing the model to capture relationships and prioritize key information more effectively.
Its scalability and parallelism make it well-suited for modeling interactions among different entities~\cite{khan2022transformers}.
Recent studies have explored Transformer-based architectures for processing complex, variable-length observations, highlighting their effectiveness in relational reasoning~\cite{zambaldi2019deep, vinyals2019grandmaster}, which is crucial for cooperative multi-agent scenarios.

    \section{Problem Formulation}\label{sec:problem-formulation}
    \subsection{Multi-Robot Multi-Target Encirclement Problem}\label{subsec:problem_formulation}
In this paper, we investigate the multi-robot multi-target encirclement problem in an uncertain environment.
The robots, including \(N\) pursuers $\{i|i=1, \ldots, N\}$ and \(M\) evaders $\{j|j=1, \ldots, M\}$, operate in a 2D bounded space with a soft boundary, meaning that exceeding the boundary is not considered a collision, but pursuers incur a penalty for leaving the designated area.
The environment includes randomly placed static obstacles and dynamic disturbances modeled using Rankine vortices~\cite{acheson1990elementary}.
The velocity of each vortex is given as
$v_r = 0$,
$v_\theta(r) = \displaystyle (\Gamma / 2\pi)
\begin{cases}
    r / r_0^2, & r \leq r_0, \\
    1 / r, & r > r_0,
\end{cases}$
where \( r \) is the distance from the vortex center, \( r_0 \) the core radius, and \(\Gamma\) the circulation.
The angular velocity of the vortex core is given by \(\Omega = \Gamma/(2\pi r_0^2)\).

Each robot is modeled as a circular entity of uniform size and follows identical kinematic constraints.
All pursuers observe their surroundings within a fixed perceptual radius and have access to evader positions.
No inter-agent communication is allowed; instead, each pursuer makes decisions independently based on local observations and a shared policy.
This distributed scheme is more robust to communication failures, and one-way target broadcasting offers a practical alternative to fragile peer-to-peer messaging.
Neither pursuers nor evaders have knowledge of the opponent's strategy.

The pursuers thrive to encircle the evaders with minimal traveling times while avoid collisions with other robots and obstacles.
We assume that a pursuer becomes inactive upon collision, whereas an encircled evader is immobilized and no longer interacts with the environment.
Pursuers that successfully encircle an evader subsequently proceed to pursue the remaining evaders until all have been encircled, satisfying the encirclement success criteria~\eqref{eq:encirclement_criteria}.
An evader is considered encircled if at least three pursuers are within the encirclement radius \( d_{\text{encircle}}\), while maintaining a maximum angular gap no greater than \( \pi \) and at most three times the minimum angular gap.
For each evader \( j \in\{1, \dots, M\}\), the encirclement criteria are defined as follows:
\begin{equation}
    \left\{
    \begin{aligned}
        &|\{\mathbf{p}_i \mid d(\mathbf{p}_i, \mathbf{p}_j) \leq d_{\text{encircle}}\}| \geq 3, \\
        &\max(\boldsymbol{\phi}) \leq \psi, \\
    \end{aligned}
    \right.
    \label{eq:encirclement_criteria}
\end{equation}
where \( \mathbf{p}_i \) and \( \mathbf{p}_j \) denote the positions of pursuer \( i \) and evader \( j \), respectively; \( d(\mathbf{p}_i, \mathbf{p}_j) \) represents their Euclidean distance; \(\psi\) controls the angular distribution.
The set \( \boldsymbol{\phi} \) consists of the angles formed between any two adjacent pursuers engaged in the encirclement and the encircled evader, where each angle is defined in the range \([0, 2\pi]\).
These constraints ensure a sufficiently uniform encirclement, preventing escape corridors and guaranteeing effective confinement.

\subsection{Reinforcement Learning}\label{subsec:marl}
The multi-target encirclement problem is formulated as a partially observable Markov decision process (POMDP).
Formally, a POMDP is defined by the tuple \( \mathcal{M} = \langle \mathcal{S}, \mathcal{A}, \mathcal{P}, \mathcal{R}, \mathcal{O}, \mathcal{Z}, \gamma \rangle \), where \( \mathcal{S} \) denotes the state space, \( \mathcal{A} \) represents the action space, \( \mathcal{P}: \mathcal{S} \times \mathcal{A} \times \mathcal{S} \to [0, 1] \) defines the state transition dynamics, \( \mathcal{R}: \mathcal{S} \times \mathcal{A} \to \mathbb{R} \) is the reward function, \( \mathcal{O} \) is the observation space, \( \mathcal{Z}: \mathcal{S} \times \mathcal{A} \times \mathcal{O} \to [0, 1] \) represents the observation probability distribution, and \( \gamma \in [0, 1] \) is the discount factor.

At each time step \( t \), the agent receives an observation \( \mathbf{o}_t \in \mathcal{O} \) and selects an action \( \mathbf{a}_t \) according to its policy \( \pi(\mathbf{a}_t | \mathbf{o}_t) \).
The environment then transits to the next state \( \mathbf{s}_{t+1} \sim \mathcal{P}(\mathbf{s}_{t+1} | \mathbf{s}_t, \mathbf{a}_t) \), and the agent receives a new observation \( \mathbf{o}_{t+1} \sim \mathcal{Z}(\mathbf{o}_{t+1} | \mathbf{s}_{t+1}, \mathbf{a}_t) \), where \( \mathbf{o}_{t+1} \in \mathcal{O} \), along with a reward \( r_t = \mathcal{R}(\mathbf{s}_t, \mathbf{a}_t) \).
The objective is to learn an optimal policy \( \pi(\mathbf{a} | \mathbf{o}) : \mathcal{O} \times \mathcal{A} \to [0,1] \) that maximizes the expected discounted cumulative reward:
\begin{equation}
    J(\pi) = \mathbb{E}_{\mathbf{a}_t \sim \pi(\cdot | \mathbf{o}_t), \mathbf{s}_{t+1} \sim \mathcal{P}} \left[ \sum_{t=0}^{\infty} \gamma^t \mathcal{R}(\mathbf{s}_t, \mathbf{a}_t) \right].
\end{equation}

\subsection{Observation and Action Spaces}\label{subsec:observation-action-spaces}
At each time step \( t \), each pursuer receives an observation \( \mathbf{o}_t \) consisting of information about itself, teammates, evaders, and obstacles within its perceptual radius \( r_\text{percept} \), denoted as \( \mathbf{o}_{\text{ego},t}, \mathbf{o}_{\text{team},t}, \mathbf{o}_{\text{evader},t}, \mathbf{o}_{\text{obstacle},t} \).
These observations are structured as:
\(
    \mathbf{o}_t = [\mathbf{o}_{\text{ego},t}, \mathbf{o}_{\text{team},t}, \mathbf{o}_{\text{evader},t}, \mathbf{o}_{\text{obstacle},t}],
\)
where
\( \mathbf{o}_{\text{ego},t} \) represents the observation of the pursuer itself,
\( \mathbf{o}_{\text{team},t} = \{\mathbf{o}_{\text{team},t}^1, \dots, \mathbf{o}_{\text{team},t}^m\} \) denotes the set of observations of teammates within the perceptual range,
\( \mathbf{o}_{\text{evader},t} = \{\mathbf{o}_{\text{evader},t}^1, \dots, \mathbf{o}_{\text{evader},t}^n\} \) denotes the set of observations of all non-encircled evaders, which can be observed regardless of their distance,
\( \mathbf{o}_{\text{obstacle},t} = \{\mathbf{o}_{\text{obstacle},t}^1, \dots, \mathbf{o}_{\text{obstacle},t}^k\} \) denotes the set of observations of obstacles within the perceptual range,
and \( m, k \) represent the number of observed teammates and obstacles within the perceptual range, while \( n \) represents the total number of evaders that are not encircled.

The observation vectors include position (\( p_x, p_y \)) for teammates, evaders, and obstacles; velocity (\( v_x, v_y \)) for teammates and evaders; and radius \( r \) for obstacles.
Spatial and directional cues in the observation include the relative distance \( d \), the egocentric angle \( \theta \) with respect to the observed entity, and the \( \text{heading\_error} \), which quantifies the evader’s orientation deviation.
The nearest obstacle distance \( d_{\text{nearest}} \) and \( \text{pursuit\_status} \) represent environmental awareness and engagement in pursuit, respectively.
\( d_{\text{nearest}} \) is set to \( r_\text{percept} \) if no obstacle is detected within the perceptual radius.
Additionally, \( \text{pursuit\_status} \) is set to 1 if any evader is within three times the encirclement threshold \( d_\text{encircle} \), and 0 otherwise.
All observations are represented in the ego-frame of the robot.

For brevity, the subscript \( t \) and superscripts indicating individual entities are omitted, and the observation vectors for each category can be expressed in compact form as
\[
    \begin{aligned}
        \mathbf{o}_{\text{ego}} &= [v_x, v_y, d_{\text{nearest}}, \text{pursuit\_status}], \\
        \mathbf{o}_{\text{team}} &= [p_x, p_y, v_x, v_y, d, \theta, \text{pursuit\_status}], \\
        \mathbf{o}_{\text{evader}} &= [p_x, p_y, v_x, v_y, d, \theta, \text{heading\_error}], \\
        \mathbf{o}_{\text{obstacle}} &= [p_x, p_y, r, d, \theta].
    \end{aligned}
\]
At each time step \( t \), given the current observation \( \mathbf{o}_t \), each pursuer independently selects an action \( \mathbf{a}_t \) according to a policy \( \pi \), such that \( \mathbf{a}_t \sim \pi(\cdot | \mathbf{o}_t) \).
The action of a pursuer is defined as \( \mathbf{a}_t = [a, \omega] \), where \( a \) denotes the linear acceleration and \( \omega \) represents the angular velocity.
In this study, a discrete action space is adopted.

    \section{Approach}\label{sec:approach}
    To handle the multi-robot setting, the centralized training with decentralized execution paradigm~\cite{gronauer2022multi} is adopted.
Global information is utilized to design the reward function, guiding robots' emergent behaviors during training and fostering collaborative and adaptive strategies.
During execution, global information is removed, and each robot independently makes decisions based on its local observations.
This approach allows robots to learn complex, adaptive strategies during training while maintaining decentralized, real-time execution capabilities.
\begin{figure*}[t!]
    \centering
    \vspace{1.5mm}
    \includegraphics[scale=0.54]{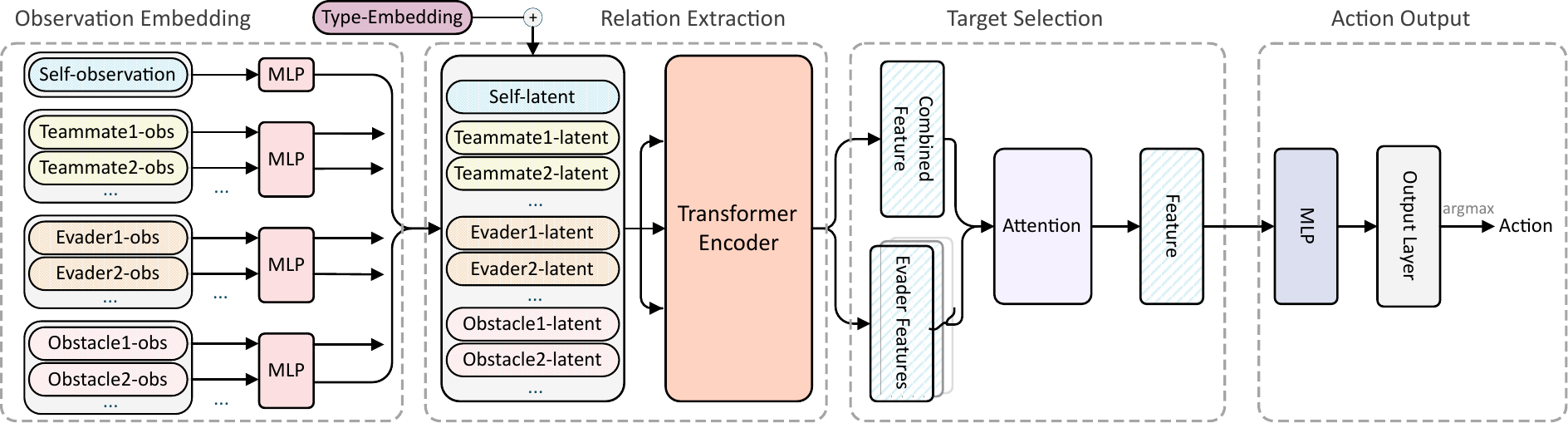} 
    \captionsetup{belowskip=-19pt}
    \caption{The network architecture of the TERL model.}
    \label{fig:terl-network-architecture}
\end{figure*}
\subsection{Reward Design}\label{subsec:reward-design}
\textbf{Safety and Guidance Reward}.
To effectively guide pursuers in tracking evaders while ensuring collision avoidance with obstacles and other robots, a distance-based reward function is designed.
This function is defined as follows:
\begin{align}
    R_{d1} &=
    \begin{cases}
        -80, & d^\text{min} < 0 \quad (\text{collision}), \\
        -5, & 0 \leq d^\text{min} < d_{\text{safe}};
    \end{cases}
    \\
    R_{d2} &=
    \begin{cases}
        5, & d_{\text{safe}} \leq d^\text{e} \leq d_{\text{encircle}}, \\
        5 \exp(-0.05 (d^e - d_{\text{encircle}})), & d^\text{e} > d_{\text{encircle}},
    \end{cases}
\end{align}
where \( d^\text{min} \) represents the minimum Euclidean distance between the pursuer and all observed teammates, evaders, and obstacles.

For each evader, a corresponding distance \( d^{e} \) is used to compute an individual reward term \( R_{d2} \) based on its proximity to the pursuer, independent of obstacles.
The overall reward \( R_{\text{d}} = R_{d1} + \sum R_{d2} \) is designed to encourage pursuers to approach evaders within \( d_{\text{encircle}} \) when feasible, thereby facilitating progressive encirclement while reducing the risk of collisions with obstacles and dynamic robots.


\textbf{Balanced target selection}.
Overcrowding is a common issue in multi-target selection, as it can hinder efficient coordination among pursuers.
To balance target selection and spatial distribution, we introduce a global reward mechanism that leverages global information to regulate cooperative behavior and ensure a more effective pursuit strategy.
We compute the number \( P \) of nearby pursuers within a specified relative distance \( d_{\text{related}} = 3 d_{\text{encircle}} \) for each evader \( j\in\{1, \ldots, M \}\).

The cooperation reward is defined as:
\begin{align}
    R_{\text{c}} =
    \begin{cases}
        -10, & \text{if } P \geq 5, \\
        5 \max\left( 0, 1.0 - 0.3 (P - 3) \right), & \text{if } 3 \leq P < 5.
    \end{cases}
\end{align}
The reward \( R_{\text{c}} \) is computed separately for each evader, and the corresponding \( P \) nearby pursuers receive either a reward or a penalty.
Additionally, if there exist both an evader with \( P < 3 \) and another with \( P > 5 \) simultaneously, all pursuers receive a penalty of \( -10 \).
Furthermore, if a pursuer crosses the arena boundary, it incurs a penalty of \( -5 \).

\textbf{Encirclement Reward.} The completion reward follows the formulation:
    \(R_{\text{completion}} = 120 \; \frac{2\pi}{n_k} e^{-\sigma}\),
where \( n_k \) represents the number of pursuers participating in the encirclement, and \( \sigma \) denotes the standard deviation of the pursuers' encirclement angles.
This formulation incentivizes a more uniform encirclement formation, enhancing coordination among pursuers.
In addition, to promote efficient task completion, a time penalty of \( -1 \) is applied at every timestep.


\subsection{TERL Policy Network Architecture}\label{subsec:transformer-enhanced-policy-network}

The TERL policy network architecture is illustrated in Fig.~\ref{fig:terl-network-architecture}.
Under the value-based reinforcement learning framework, Implicit Quantile Networks (IQN)~\cite{dabney2018implicit} are employed to estimate the action-value distribution.
The TERL model consists of four sequential components: \textit{observation embedding}, \textit{relation extraction}, \textit{target selection}, and \textit{action output}.

\paragraph{Observation Embedding}
This module encodes raw observations into latent representations.
As defined in Subsection~\ref{subsec:observation-action-spaces}, each teammate, obstacle, and evader is structured as an observation vector.
Unlike approaches that concatenate all observations into a single input, we keep the observation of each entity as an independent input.
To preserve the distinct characteristics of different entity types, we assign each type a dedicated embedding layer, enabling the model to capture type-specific representations and prevent information mixing across categories.
As illustrated in Fig.~\ref{fig:terl-network-architecture}, each entity type is processed by a respective multi-layer perceptron (MLP), which maps it to a latent vector of uniform dimension.
This module outputs a set of latent vectors \( E \) with shape \( (\ell, f) \), where \( \ell \) denotes the number of entities and \( f \) represents the latent feature dimension.
Each row of \( E \) corresponds to the latent representation of an individual entity, which is then fed into the \textit{Relation Extraction} module.

\paragraph{Relation Extraction}
This module captures entity interactions using a pairwise attention mechanism.
To incorporate entity type information, we introduce type embeddings into the input set \( E \), where each entity—ego, teammates, obstacles, and evaders—is assigned a type identifier \( t \in \{0, 1, 2, 3\} \).
Each identifier is mapped to a continuous representation \( \mathbf{T} \in \mathbb{R}^{f} \), ensuring type-specific feature encoding while maintaining consistency with the observation embedding space.
The resulting type embedding \( \mathbf{T} \) is added to the corresponding entity’s latent vector, allowing the model to differentiate interaction patterns across different entity types.

For each entity \( i \), interactions are computed with all other entities, including itself, leading to an updated representation:
\(
s_i = \sum_{j=1}^{\ell} \mathbf{r}_{i,j},
\)
where \( \mathbf{r}_{i,j} \) represents the interaction between entities \( i \) and \( j \).
To model these interactions, we employ a self-attention mechanism within the transformer framework, enabling each pursuer to dynamically integrate information from teammates, evaders, and obstacles.
This is particularly crucial in pursuit-evasion tasks, where pursuers must adaptively assess their surroundings to make informed decisions.

In our architecture, the latent representations of the pursuer and observed entities are projected into query \( q \), key \( k \), and value \( v \), enabling structured feature interactions across entities.
The scaled dot-product attention quantifies the relevance between the pursuer’s query and the keys of all observed entities, assigning higher weights to more relevant interactions.
Mathematically, the interaction between the pursuer and another entity is defined as:
\begin{equation}
    \mathbf{r}_{0,j} = \text{softmax} \left( \frac{q_0 k_j^T}{\sqrt{d_k}} \right) v_j,
    \label{eq:interaction}
\end{equation}
where \( q_0 \) represents the pursuer’s query, \( d_k \) is the dimension of the key, and \( k_j, v_j \) are the key and value of entity \( j \) (including the pursuer itself).
This mechanism allows the pursuer to prioritize relevant entities in its surroundings, focusing on key aspects such as evaders’ movements, teammate coordination, and obstacle constraints.

The resulting attention weights \( w_{0,j} = \text{softmax} ( \frac{q_0 k_j^T}{\sqrt{d_k}} ) \) reflect the relative importance of each entity in the pursuer’s decision-making process.
By aggregating weighted value vectors, the latent state of pursuer is updated as:
\(
s_0 = \sum_{j=1}^{\ell} w_{0,j} v_j.
\)
This formulation enables the model to capture key pursuit-evasion dynamics:
1) identifying and prioritizing appropriate targets,
2) coordinating with teammates to enhance encirclement efficiency, and
3) dynamically adjusting movement strategies to navigate around obstacles.
Thus, self-attention serves as an adaptive information fusion process, improving each pursuer’s situational awareness and decision-making.
By stacking all queries, keys, and values into matrices \( Q \), \( K \), and \( V \), the pairwise interactions can be efficiently computed in matrix form:
\(
S = \text{softmax} \left( \frac{QK^T}{\sqrt{d_k}} \right) V.
\)

The multi-head attention mechanism further enhances interaction modeling by capturing diverse relational patterns, ensuring a more robust decision-making.
Each attention head captures distinct interaction features, and their outputs are concatenated to produce the final entity representation \( M \), maintaining the same shape as the input \( E \).
This iterative process enhances relational modeling, enabling the model to refine its understanding of pursuit-evasion dynamics.

\paragraph{Target Selection}
To extract meaningful features for \textit{target selection}, we first apply max-pooling over \( M \) to obtain a compact global feature that captures key interaction features across all entities.
The ego's latent representation in \( M \) is then concatenated with this global feature to form the combined feature, integrating both local and global context.
Meanwhile, evader representations are extracted separately to preserve essential target-specific information.

The \textit{target selection} module transforms the combined feature into a query, while evader representations serve as keys and values, all with dimension \( f \).
The computation follows the same attention mechanism as in~\eqref{eq:interaction}, dynamically adjusting focus based on task-critical factors such as proximity and movement patterns.
This ensures pursuers prioritize the most relevant evaders for effective decision-making.
The weighted sum is concatenated with the query to form the final representation, which is then passed to the action output module.
By leveraging structured attention, this mechanism adaptively refines focus, enhancing coordination and decision-making in multi-agent interactions.

\paragraph{Action Output}
Based on the output of the \textit{target selection} module, the action-value distribution is estimated using the IQN algorithm.
The final action is selected via the argmax operation over the expected value distribution.

\subsection{TERL Policy Training}\label{subsec:terl-policy-training}

Building upon the TERL policy network architecture, we employ curriculum learning and random initialization techniques to train the TERL model.
Curriculum learning improves model performance by progressively introducing training examples from simple to complex, facilitating convergence and generalization~\cite{bengio2009curriculum}.
In the proposed training framework, curriculum learning gradually increases environment difficulty by adjusting configuration parameters at predefined time steps, enhancing robot adaptability and robustness.

    \section{Experiments}\label{sec:experiments}
    
\subsection{Training Setup}\label{subsec:training-setup}
The simulation environment is adapted from the multi-robot navigation simulator~\cite{lin2024decentralized}, specifically tailored for the multi-target encirclement scenario.
To account for the limited perception range of each pursuer, the maximum number of detectable obstacles and teammates is set to 5, while the maximum number of observable evaders is limited to 8, but can be adjusted during evaluation to assess performance under different scenarios.
All detected entities are sorted based on their distance to the pursuer.
If the number of observed entities is below the predefined threshold, zero-padding is applied, and the corresponding positions in the masking vector are set to zero.
To promote a uniform encirclement distribution, the criterion \(\max(\boldsymbol{\phi}) \leq \kappa \min(\boldsymbol{\phi})\) is included in~\eqref{eq:encirclement_criteria}, with \(\psi\) and \(\kappa\) set to \(\pi\) and 3, respectively.
Each episode has a maximum duration of 3000 timesteps and terminates under the following conditions: (i) all evaders are successfully encircled, (ii) the number of active pursuers falls below three, or (iii) the timestep limit is reached.
At the beginning of each episode, the environment is reset, with pursuers, evaders, static obstacles, and vortices randomly initialized according to the curriculum training schedule, as detailed in \autoref{tab:curriculum_schedule}.
\begin{table}[hp]
    \vspace{-1.7mm}
    \centering
    \caption{\textbf{Curriculum Training Schedule}. Numbers indicate the count of each entity type.}
    \label{tab:curriculum_schedule}
    \renewcommand{\arraystretch}{1.2}
    \setlength{\tabcolsep}{5.5pt}
    \begin{tabular}{c c c c c}
        \toprule
        \textbf{Time Step (Million)} & \textbf{Pursuers} & \textbf{Evaders} & \textbf{Obstacles} & \textbf{Vortices} \\
        \midrule
        \( 0 \leq t < 2 \)           & 3                 & 1                & 0                  & 4                 \\
        \( 2 \leq t < 4 \)           & 4                 & 1                & 1                  & 6                 \\
        \( 4 \leq t < 5 \)           & 7                 & 2                & 2                  & 8                 \\
        \( 5 \leq t < 6 \)           & 11                & 3                & 4                  & 8                 \\
        \( 6 \leq t \leq 7 \)        & 15                & 4                & 6                  & 8                 \\
        \bottomrule
    \end{tabular}
    \vspace{-6mm}
\end{table}
\subsection{Evaluation Setup}\label{subsec:evaluation-setup}
To comprehensively evaluate the performance of the proposed TERL framework, we conducted extensive experiments comparing it with the baseline methods introduced in Subsection~\ref{subsec:baseline-methods}.
Additionally, ablation studies were performed to assess the contributions of the \textit{Relation Extraction} and \textit{Target Selection} modules within the TERL framework.

To ensure a fair comparison, all methods were trained and evaluated under identical training (all using curriculum learning) and testing conditions.
\autoref{tab:evaluation_schedule} summarizes the experimental scenario configurations.
In standard scenarios (small, medium, and large scale), the number of pursuers is maintained at an approximate 3–4:1 ratio relative to the number of evaders.
In contrast, the constrained coordination (CC) configuration enforces a stricter 2:1 pursuer-to-evader ratio, significantly increasing the demand for cooperation, strategic task allocation, and efficient decision-making.
Each scenario consists of 20 independent experiments, where robots, obstacles, and vortices are randomly initialized at the beginning of each trial.
The maximum episode length is set to 1000 timesteps.

The pursuers select acceleration values from the discrete set \( a \in \{-0.4, 0.0, 0.4\} \) (m/s\(^2\)) and angular velocity values from \( \omega \in \{-\pi/6, 0, \pi/6\} \) (rad/s), with a maximum speed constraint of 3.0 m/s.
Evaders employ an artificial potential field (APF) strategy~\cite{khatibRealtimeObstacle1986} to evade capture.
They possess a higher maximum speed of 3.5 m/s and a broader range of angular velocity options, given by \( \omega \in \{-\pi/6, -\pi/12, 0, \pi/12, \pi/6\} \) (rad/s), enabling enhanced maneuverability and more agile evasion strategies.

We evaluate the methods based on three key performance metrics: \textbf{success rate}, \textbf{travel time}, and \textbf{collision ratio}.
A success in an experiment means that all evaders are encircled within the timestep limit.
The success rate represents the proportion of successful experiments, indicating the method's effectiveness in achieving the encirclement objective.
Travel time is defined as the timestep when each episode terminates under the conditions specified in Subsection~\ref{subsec:training-setup}, reflecting the method’s ability to coordinate and sustain pursuit until completion or failure.
The collision ratio quantifies the fraction of pursuers that collide with obstacles or other pursuers, where a lower value suggests safer and more efficient navigation.
\begin{table}[t]
    \centering
    \vspace{1.3mm}
    \caption{\textbf{Experimental Scenario Configurations}. The values indicate the number of entities in each scenario.}
    \label{tab:evaluation_schedule}
    \renewcommand{\arraystretch}{1.2}
    \setlength{\tabcolsep}{7.8pt}
    \begin{tabular}{l c c c c}
        \toprule
        \textbf{Scenario} & \textbf{Pursuers} & \textbf{Evaders} & \textbf{Obstacles} & \textbf{Vortices} \\
        \midrule
        \multirow{3}{*}{Small Scale}
        & 11                & 3                & 2                  & 8                 \\
        & 15                & 4                & 4                  & 8                 \\
        & 19                & 5                & 6                  & 8                 \\
        \midrule
        \multirow{3}{*}{Medium Scale}
        & 48                & 12               & 8                  & 8                 \\
        & 51                & 13               & 8                  & 8                 \\
        & 56                & 14               & 8                  & 8                 \\
        \midrule
        \multirow{3}{*}{Large Scale}
        & 72                & 18               & 8                  & 8                 \\
        & 76                & 19               & 8                  & 8                 \\
        & 80                & 20               & 8                  & 8                 \\
        \midrule
        {CC}
        & 28                & 14               & 8                  & 8                 \\
        \bottomrule
    \end{tabular}
    \vspace{-6mm}
\end{table}

\begin{figure*}[t!]
    \centering
    \vspace{1.5mm}
    \includegraphics[scale=0.88]{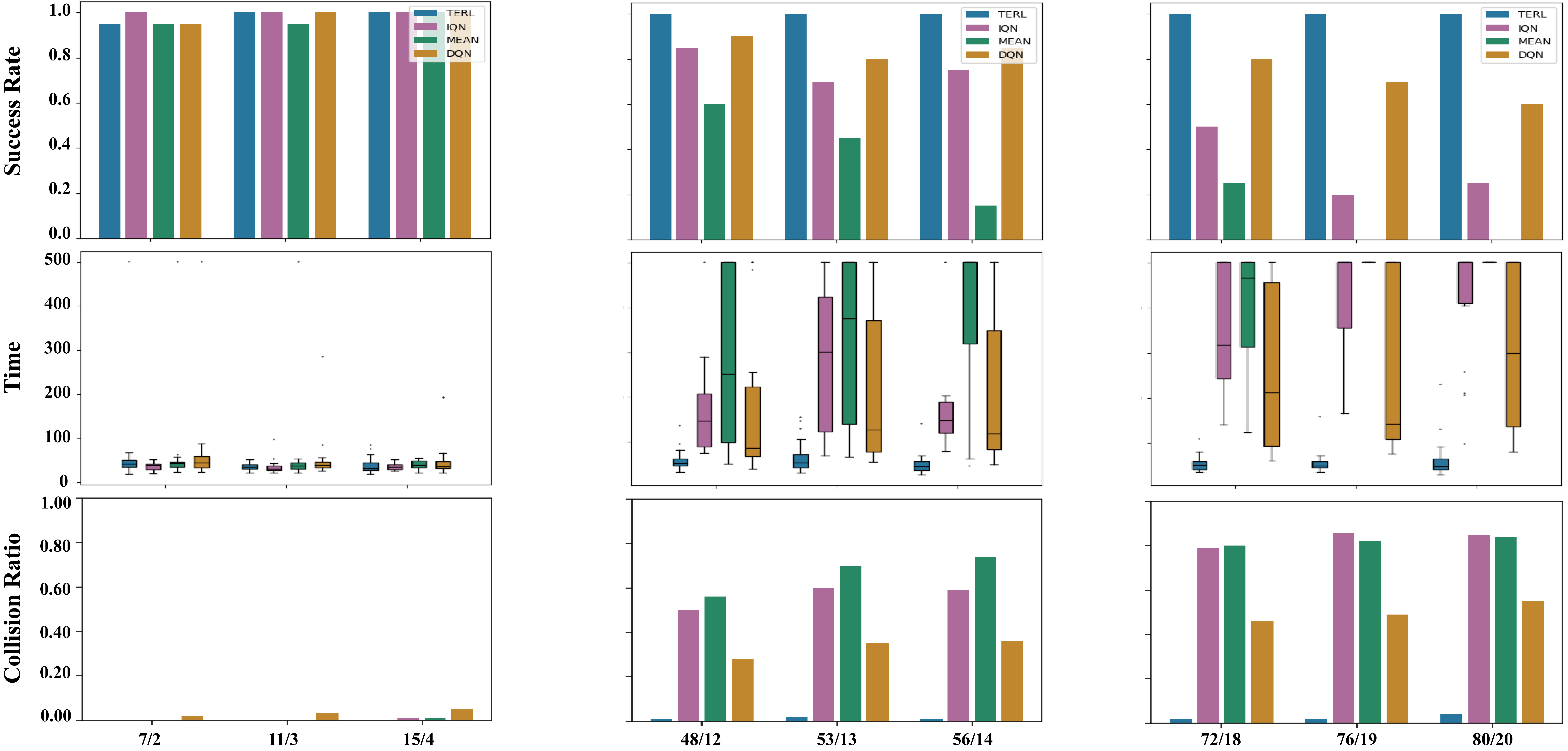} 
    \caption{\textbf{Experimental results.} The top, middle, and bottom rows show success rates, travel times, and collision ratios for small, medium, and large scales. The horizontal axis represents the number of pursuers and evaders (pursuer/evader).}
    \label{fig:exp_results}
    \vspace{-5mm}
\end{figure*}
\subsection{Baseline Methods}\label{subsec:baseline-methods}
We compare the proposed TERL framework with three baselines: IQN~\cite{dabney2018implicit}, DQN~\cite{mnih2015human}, and Mean Embedding.
In all three baselines, the \textit{Relations Extraction} and \textit{Target Selection} modules are removed.
In IQN and DQN, average pooling is applied to the \textit{Observation Embedding} outputs within each entity category, followed by concatenation with the Action Output module.
In the Mean Embedding baseline, \textit{Observation Embedding} outputs are averaged within each category, and the resulting mean features are then averaged across categories to form a single global feature vector, which is fed into the Action Output module of the IQN framework.
By comparing these algorithms, we aim to evaluate the role of relational reasoning in multi-target encirclement.

\subsection{Results and Discussion}\label{subsec:results-and-discussion}

The experimental results are summarized in Fig.~\ref{fig:exp_results} and \autoref{tab:specical_config_results}.
We evaluate performance based on three key metrics: success rate, traveling time (measured in seconds), and collision ratio.
\begin{table}[h]
    \centering
    \vspace{-1.6mm}
    \caption{\textbf{CC Configuration Results}}
    \label{tab:specical_config_results}
    \renewcommand{\arraystretch}{1.2} 
    \setlength{\tabcolsep}{10pt} 
    \begin{tabular}{c c c c c}
        \toprule
        \textbf{Method} & \textbf{Success Rate} & \textbf{Time} & \textbf{Collision Ratio} \\
        \midrule
        \text{TERL}     & \textbf{0.80}                  &226.78        &\textbf{0.03}                     \\
        \text{IQN}      & 0.00                  & \textbf{150.50}        & 0.93                     \\
        \text{MEAN}     & 0.00                  & 210.75        & 0.92                     \\
        \text{DQN}      & 0.00                  & 252.45        & 0.90                     \\
        \bottomrule
    \end{tabular}
    \vspace{-4mm}
\end{table}
\subsubsection{Completion}
TERL consistently outperforms all baselines across all scenarios. In small-scale settings, all methods achieve over 95\% success, indicating that traditional feature aggregation suffices for simple cases. However, as the scale increases, IQN and DQN experience significant performance degradation, while Mean Embedding performs the worst. TERL maintains a 100\% success rate, demonstrating superior scalability. The CC scenario further highlights this advantage, where TERL achieves 80\% success, whereas all baselines fail completely (0\%). This underscores TERL’s adaptive relational reasoning, which enables effective coordination under constrained conditions.

\subsubsection{Optimality of Executions}
TERL consistently achieves lower and more stable travel times compared to the baselines.
In small-scale scenarios, IQN exhibits slightly shorter travel times; however, as task complexity increases, TERL surpasses all methods, demonstrating superior decision-making efficiency.
The steady variance and mean of TERL indicate its robustness and reliable performance as the team scales up, in contrast to the increasing variance of IQN and DQN and the frequent failures of Mean Embedding.
In the CC scenario, TERL successfully completes encirclement in 226.78 seconds, whereas all baselines fail entirely, further highlighting the advantages of relation-aware coordination.

\subsubsection{Safety}
TERL achieves the lowest collision ratio across all scenarios.
As task complexity increases, the baselines exhibit a significant rise in collision rates.
Meanwhile, TERL maintains a consistently low collision rate.
In large-scale scenarios, frequent collisions hinder successful encirclement for all baselines.
The CC scenario presents the most striking contrast: TERL sustains a collision ratio of 0.03, while all baselines exceed 90\%, rendering them ineffective.

\subsubsection{Overall Analysis}
The performance degradation of IQN, DQN, and Mean Embedding with increasing scale highlights the limitations of basic feature aggregation. Mean Embedding loses entity-specific information, while IQN and DQN struggle with scalability.
In contrast, TERL consistently excels across all metrics due to:
1) Relational Reasoning for effective multi-agent coordination.
2) Adaptive Target Selection for efficient pursuit.
3) Scalability in handling large-scale, complex environments.
These results demonstrate that TERL is well-suited for multi-target encirclement, outperforming baselines in robustness, efficiency, and safety.

Moreover, the generalizability of TERL arises from its capability to model inter-agent dependencies in a decoupled and relational manner.
Instead of aggregating inputs at early layers, TERL retains individual entity embeddings and applies attention mechanisms to directly capture transferable interaction patterns.
This relational modeling allows the policy to remain sensitive to structural cues such as relative distances, spatial configurations, and target-specific interactions.
Consequently, TERL naturally adapts to varying numbers of entities and interaction complexities, enabling robust performance in large-scale and dynamic scenarios.



\subsection{Ablation Study}\label{subsec:ablation-study}
This ablation study evaluates the impact of the \textit{Relation Extraction} and \textit{Target Selection} modules on TERL’s performance.
Removing Relation Extraction replaces the Transformer module with IQN’s aggregation mechanism, while removing Target Selection feeds the output of Relation Extraction directly into the Action Output module.
The results in \autoref{tab:ablation_results} highlight the necessity of these components in maintaining TERL’s high success rate, low travel time, and minimal collisions.
\begin{table}[t]
    \caption{\textbf{Ablation Study Results}}
    \label{tab:ablation_results}
    \centering
    \renewcommand{\arraystretch}{1.2}
    \setlength{\tabcolsep}{5pt}
    \begin{tabular}{l p{2.8cm} >{\centering\arraybackslash}p{1cm} >{\centering\arraybackslash}p{1cm} >{\centering\arraybackslash}p{1cm}}
        \toprule
        \multirow{2}{*}{\textbf{Scenario}} & \multirow{2}{*}{\textbf{Method}}
        & \textbf{Success Rate} & \multirow{2}{*}{\textbf{Time}} & \textbf{Collision Ratio} \\
        \midrule
        \multirow{3}{*}{Small}
        & TERL                    & \textbf{1.00} & 45.98           & \textbf{0.00} \\
        & w/o Relation-Extraction & \textbf{1.00} & \textbf{36.39}  & 0.02          \\
        & w/o Target-Selection    & \textbf{1.00} & 52.98           & \textbf{0.00} \\
        \midrule
        \multirow{3}{*}{Medium}
        & TERL                    & \textbf{1.00} & \textbf{63.80}  & \textbf{0.01} \\
        & w/o Relation-Extraction & \textbf{1.00} & 81.48           & 0.45          \\
        & w/o Target-Selection    & \textbf{1.00} & 78.02           & 0.03          \\
        \midrule
        \multirow{3}{*}{Large}
        & TERL                    & \textbf{1.00} & \textbf{62.62}  & \textbf{0.03} \\
        & w/o Relation-Extraction & 0.75          & 161.94          & 0.75          \\
        & w/o Target-Selection    & \textbf{1.00} & 84.68           & 0.04          \\
        \midrule
        \multirow{3}{*}{CC}
        & TERL                    & \textbf{1.00} & 194.05 & \textbf{0.02} \\
        & w/o Relation-Extraction & 0.05          & \textbf{100.42}          & 0.91          \\
        & w/o Target-Selection    & 0.55          & 390.60          & 0.06          \\
        \bottomrule
    \end{tabular}
    \vspace{-8mm}
\end{table}

\textbf{Impact of \textit{Relation Extraction} (RE).} Removing RE significantly degrades performance in large and complex scenarios.
Although success rates remain unaffected in small and medium scales, travel time increases (e.g., from 63.80 to 81.48 in medium scale).
In the large-scale setting, the success rate drops from 1.00 to 0.75, while travel time and collision ratio surge from 62.62 to 161.94 and 0.03 to 0.75, respectively.
The CC scenario exhibits the most severe impact, with the success rate plummeting to 0.05 and the collision ratio rising to 0.91.
Overall, the absence of RE makes the policy more aggressive and poorly coordinated; while faster in some simple tasks, this results in overall poor performance, underscoring the necessity of RE for effective coordination in diverse scenarios.

\textbf{Impact of \textit{Target Selection}.} While the absence of \textit{Target Selection} mainly affects coordination efficiency, it also leads to a decline in success rates in highly constrained scenarios.
Across standard scenarios, travel time increases consistently (e.g., from 63.80 to 78.02 in medium scale and from 62.62 to 84.68 in large scale), indicating reduced coordination efficiency.
The CC scenario reveals a more pronounced effect, with the success rate dropping to 0.55 and travel time nearly doubling to 390.60, highlighting inefficiencies in task execution without explicit target selection.

These results underscore the critical role of both modules: \textit{Relation Extraction} ensures robust multi-robot coordination, particularly in large and constrained environments, while \textit{Target Selection} improves task efficiency by optimizing action assignment. Together, they enable TERL to achieve superior performance in multi-target encirclement tasks.

    \section{Conclusion}\label{sec:conclusion}
    This paper introduces TERL, a Transformer-enhanced reinforcement learning framework for large-scale multi-target encirclement.
By incorporating relation extraction and target selection mechanisms, TERL effectively models inter-agent interactions and prioritizes targets, achieving higher success rates, shorter travel times, and lower collision ratios across different scales.
Extensive experiments validate TERL’s robustness and scalability.
The ablation study underscores the essential role of relation extraction in ensuring task success and the impact of target selection on execution efficiency.

    \FloatBarrier  
    \bibliographystyle{IEEEtran}

\end{document}